\documentclass[10pt,twocolumn,letterpaper]{article}

\usepackage{cvpr}
\usepackage{times}
\usepackage{epsfig}
\usepackage{graphicx}
\usepackage{amsmath}
\usepackage{amssymb}
\usepackage{multirow}
\usepackage{array}

\usepackage{algorithm} 
\usepackage{algpseudocode} 

\usepackage{authblk}
\makeatletter
\renewcommand\AB@affilsepx{\quad\protect\Affilfont} 
\makeatother

\graphicspath{{Imgs/}}

\usepackage[pagebackref=true,breaklinks=true,letterpaper=true,colorlinks,bookmarks=false]{hyperref}

 \cvprfinalcopy 

\def\cvprPaperID{4764} 

\begin{document}

\title{Multiple Object Tracking by Flowing and Fusing}


\author{Jimuyang Zhang~\textsuperscript{1}\thanks{Contributed equally.}, Sanping Zhou~\textsuperscript{2}$^*$, Xin Chang~\textsuperscript{3}, Fangbin Wan~\textsuperscript{4}, \\Jinjun Wang~\textsuperscript{2}, Yang Wu~\textsuperscript{5}, Dong Huang~\textsuperscript{1}
}
\affil{
\textsuperscript{1}Carnegie Mellon University,~\textsuperscript{2}Xi'an Jiaotong University,\\~\textsuperscript{3}Nara Institute of Science and Technology,~\textsuperscript{4}Fudan University,~\textsuperscript{5}Kyoto University
} 

\maketitle
\begin{abstract}
Most of Multiple Object Tracking~(MOT) approaches compute individual target features for two subtasks: estimating target-wise motions and conducting pair-wise Re-Identification~(Re-ID). Because of the indefinite number of targets among video frames, both subtasks are very difficult to scale up efficiently in end-to-end Deep Neural Networks~(DNNs). In this paper, we design an end-to-end DNN tracking approach, Flow-Fuse-Tracker~(FFT), that addresses the above issues with two efficient techniques: target flowing and target fusing. Specifically, in target flowing, a FlowTracker DNN module learns the indefinite number of target-wise motions jointly from pixel-level optical flows. In target fusing, a FuseTracker DNN module refines and fuses targets proposed by FlowTracker and frame-wise object detection, instead of trusting either of the two inaccurate sources of target proposal. Because FlowTracker can explore complex target-wise motion patterns and FuseTracker can refine and fuse targets from FlowTracker and detectors, our approach can achieve the state-of-the-art results on several MOT benchmarks. As an online MOT approach, FFT produced the top MOTA of 46.3 on the 2DMOT15, 56.5 on the MOT16, and 56.5 on the MOT17 tracking benchmarks, surpassing all the online and offline methods in existing publications.~\footnote{We will open source our code soon after the review process.} 
\end{abstract}

\section{Introduction}
\label{sec:intro}
\label{sec:intro}
Multiple Object Tracking~(MOT)~\cite{Yoon_Boragule_Song:2018,Milan_Rezatofighi_Dick:2017} is a critical perception technique required in many computer vision applications, such as autonomous driving~\cite{Chen_Seff_Koenhauser:2015}, field robotic~\cite{Ross_English_Ball:2015} and video surveillance~\cite{Zhou_Wang_Wang:2017}. One of the most popular MOT settings is tracking-by-detection~\cite{Zhang_Wang_Wang:2015}, in which the image-based object detection results are available in each frame, and MOT algorithms basically associate the same target between frames. Because of occlusion, diversity of motion patterns and the lack of tracking training data, MOT under the tracking-by-detection setting remains a very challenging task in computer vision.

\begin{figure}[t]
	\centering
	\includegraphics[width=1\linewidth]{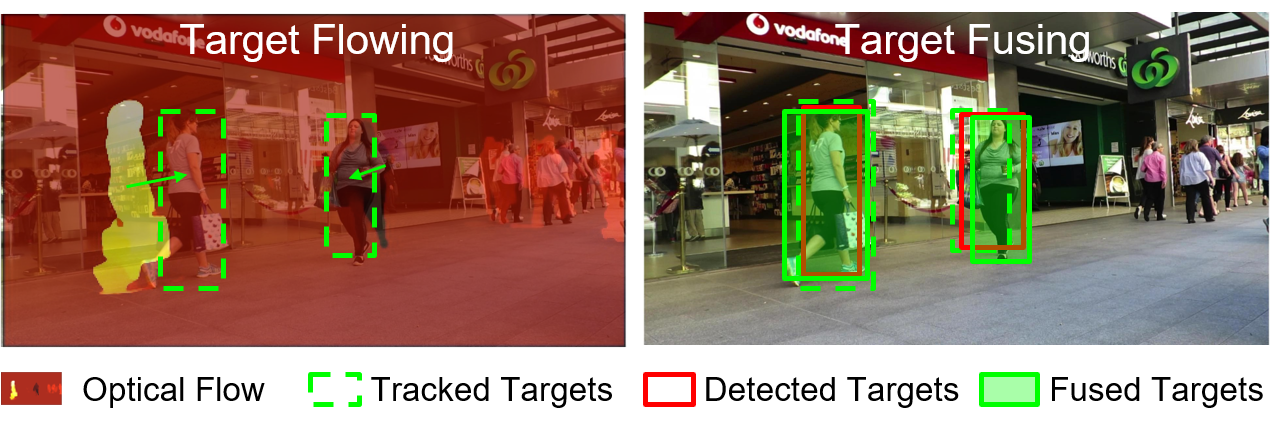}
	\caption{Multiple object tracking by target flowing and target fusing. \textbf{Target Flowing}: a FlowTracker DNN module learns indefinite number of target-wise motion jointly from frame-wise optical flows. \textbf{Target Fusing}: a FuseTracker DNN module jointly refines and fuses targets from FlowTracker and image based object detection, instead of trusting either of the two inaccurate sources of target proposal. (\textbf{Best viewed in color}) }
	\label{fig:FFTfirst}
\end{figure}

Recent MOT approaches~\cite{henschel2018fusion,Bae_Yoon:2018} use Deep Neural Networks~(DNNs) to model individual target motions and compare pairwise appearances for target association. Benefiting from the powerful representation capability of DNNs, both the target-wise features for motion and pair-wise features for appearance have been significantly improved to boost the MOT performance. However, the computation of both features is very difficult to scale up efficiently for indefinite number of targets among video frames. For instance, to compare pair-wise appearances among indefinite number of targets, one needs to process indefinite number of pairs iteratively using fixed-structured DNNs. 

In practice, most approaches solve the MOT problem in two separated steps: 1) Run a motion model and an appearance model on input frames to produce motion and appearance features, respectively; and 2) Conduct target association between frames based on the motion and appearance features. In general, the target association step finds one-to-one matches using the Hungarian algorithm~\cite{Kuhn:1955}. However, because the two steps are not optimized in an end-to-end framework, a heavy DNN is usually needed to learn highly discriminative motion and appearance features. Without nearly perfect features, numerous candidate target pairs and heuristic tricks need to be processed in the target association step. Therefore, it is very hard to develop an efficient MOT system when the feature extraction and data association are treated as two separate modules. 



In this paper, we design an end-to-end DNN tracking approach, Flow-Fuse-Tracker~(FFT), that addresses the above issues with two efficient techniques: target flowing and target fusing.  The end-to-end nature of FFT significantly boosts the MOT performance even without using the time-consuming pairwise appearance comparison. As shown in Figure~\ref{fig:FFTfirst}, FFT applies target flowing, a FlowTracker DNN module, to extract multiple target-wise motions from pixel-level optical flows. While in target fusing, a FuseTracker DNN module refines and fuses targets proposed by FlowTracker and frame-wise object detection, instead of trusting either of the two inaccurate sources of target proposal. In the training process, we simply take two regression losses and one classification loss to update the weights of both FlowTracker and FuseTracker DNN modules. As a result, the FlowTracker learns complex target-wise motion patterns from simple pixel-level motion clues, and the FuseTracker gains the ability to further refine and fuse inaccurate targets from FlowTracker and object detectors. Given paired frames as input, FFT will directly output the target association results in the tracking process.


In summary, the main contributions of our work can be highlighted as follows:
\begin{itemize}
	\item An end-to-end DNN tracking approach, called FFT, that jointly learns both target motions and associations for MOT.  
	\item A FlowTracker DNN module that jointly learns indefinite number of target-wise motions from pixel-level optical flows.
	\item A FuseTracker DNN module that directly associates the indefinite number of targets from both tracking and detection.
	\item As an online MOT method, FFT achieves the state-of-the-art results among all the existing online and offline approaches.
\end{itemize}

\section{Related Work}
Because most of the existing MOT methods consist of a motion model and an appearance model, we review variants of each model, respectively.

\textbf{Motion Models}. The motion models try to describe how each target moves from frame to frame, which is the key to choose an optimal search region for each target in the next frame. The existing motion models can be roughly divided into two categories, \emph{i,e.}, single-target models and multi-target models. In particular, the single-object models usually apply the linear methods~\cite{Kalman:1960, Milan_Schindler_Roth:2016,Oron_Bar_Avidan:2014,Zhu_Yang_Liu:2018} or nonlinear methods~\cite{Dicle_Camps_Sznaier:2013,Yang_Nevatia:2012} to estimate the motion pattern of each object across adjacent frames. In the earlier works, the Kalman filter~\cite{Kalman:1960} has been widely used to estimate motions in many MOT methods~\cite{Andriyenko_Schindler:2011,Andriyenko_Schindler_Roth:2012,Wojke_Bewley_Paulus:2017}. In recent years, a large number of deep learning based methods are designed to model motion patterns given sequences of target-wise locations. For example, the Long Short Term Memory~(LSTM) networks~\cite{Milan_Rezatofighi_Dick:2017,Sadeghian_Alahi_Savarese:2017,Kim_Li_Rehg:2018} describe and predict complex motion patterns of each target over multiple frames. Different from single-target approaches, the multi-target models aim to estimate the motion patterns of all the targets in each frame, simultaneously. Representative methods take optical flows~\cite{xie2019object, fragkiadaki2012two, keuper2018motion}, LiDAR point clouds~\cite{wu2018squeezeseg, kampker2018towards}, or stereo point clouds~\cite{mitzel2012taking, bajracharya2009fast} as inputs to estimate different motion patterns.  All these approaches use frame-wise point/pixel-level motion as a feature to match individual objects. Their matching processes are clustering in nature, which leverages multiple features such as trajectory and appearances. In comparison, our FlowTracker is a simple regression module that directly estimates target-wise motion from the category-agnostic optical flow.

\begin{figure*}[t]
\centering
\includegraphics[width=\linewidth]{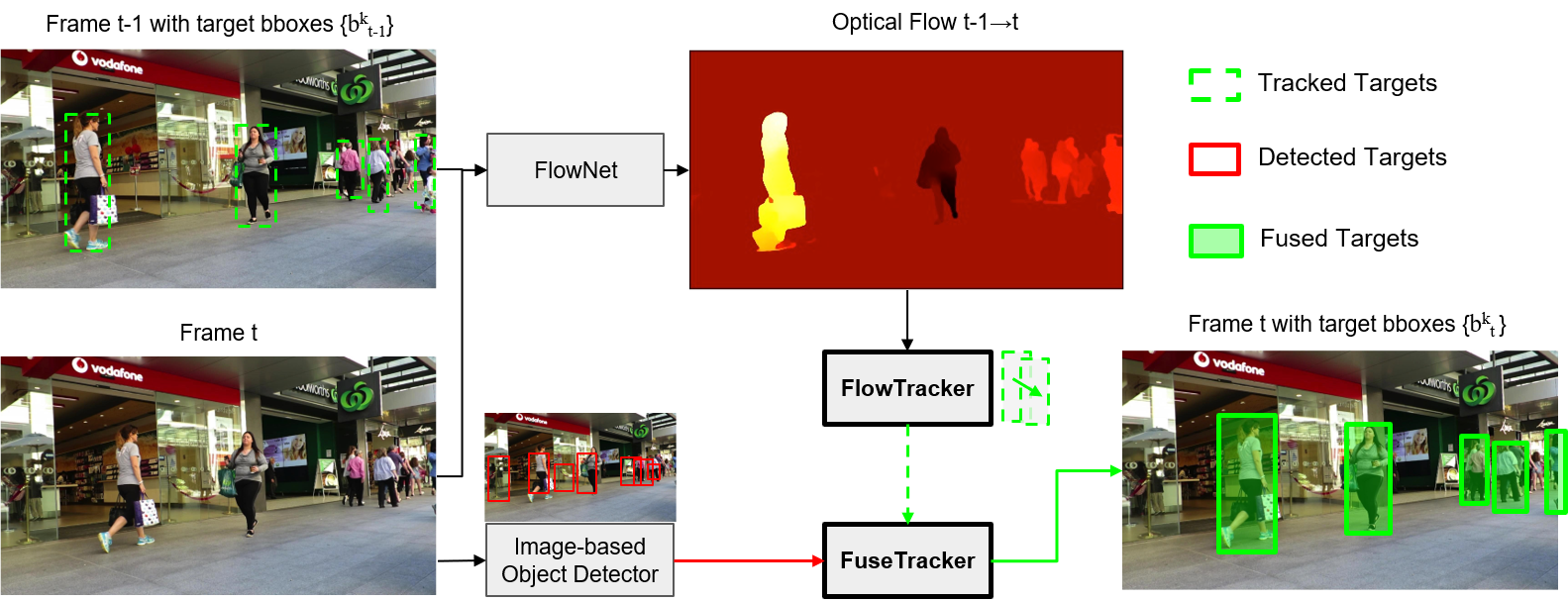}\hspace{-.0cm}
\caption{The overall view of Flow-Fuse-Tracker~(FFT) for multiple object tracking. FlowTracker and FuseTracker (in bold grey boxes) are the two DNN modules of our FFT network. In the FlowTracker, the optical flow is generated from two sequential frames, and the target bboxes $\{b^k_{t-1}\}$~(\textcolor{green}{green} dashed bboxes) at frame $t-1$ are regressed to the bboxes $\{b^k_t\}$ at frame $t$ through the optical flow. In the FuseTracker, the bboxes from both $\{b^k_t\}$ and public detections $D_t$~(\textcolor{red}{red} bboxes) at frame $t$ are refined and fused. The FuseTracker outputs the final tracking results~(\textcolor{green}{green} bboxes with shadow). (\textbf{Best viewed in color})}
\label{FigFFTNetwork}
\end{figure*}

\textbf{Appearance Model}. The appearance models aim to learn discriminative features of targets, such that features of the same targets across frames are more similar than those of different targets. In the earlier tracking approaches, the color histogram~\cite{Choi_Savarese:2010,Leibe_Schindler_Cornelis:2008,Le_Heili_Odobez:2016} and pixel-based template representations~\cite{Wu_Thangali_Sclaroff:2012,Pellegrini_Ess_Schindler:2009} are standard hand-crafted features used to describe the appearance of moving objects. In addition to using Euclidean distances to measure appearance similarities, the covariance matrix representation is also applied to compare the pixel-wise similarities, such as the SIFT-like features~\cite{Fulkerson_Vedaldi_Soatto:2008,Low:2004} and pose features~\cite{Roth_Nevatia_Stiefelhagen:2012,Qin_Shelton:2016}. In recent years, the deep feature learning based methods have been popularized with the blooming of DNNs.  For example, the features of multiple convolution layers are explored to enhance the discriminative capability of resulting features~\cite{Chen_Ai_Shang:2017}. In~\cite{Bae_Yoon:2018}, an online feature learning model is designed to associate both the detection results and short tracklets.  Moreover, different network structures, such as siamese network~\cite{Leal_Canton-Ferrer_Schindler:2016}, triplet network~\cite{Ristani_Tomasi:2018} and quadruplet network~\cite{Son_Baek_Cho:2017}, have been extensively used to learn the discriminative features cropped from the detected object bounding boxes. Benefiting from the powerful representation capability of DNNs, this line of methods~\cite{Wan_Wang_Kong:2018,Sun_Akhtar_Song:2018,Fernando_Denman_Sridharan:2018} have achieved very promising results on the public benchmark datasets. However, most of these approaches extract discriminative features independently for each object bounding box, which is very costly when scaling-up to indefinite number of objects in MOT settings.

\section{Our Method}
To achieve an end-to-end MOT DNN, the fundamental demands are to jointly process the indefinite number of targets in both motion estimation and target association.  

We address these demands with two techniques: 1) Target Flowing. Instead of extracting the target-wise features from images, our FlowTracker first estimates the motions of all pixels by optical flow and then refines the motion for each target. This technique shares the pixel-level appearance matching of optical flow over all targets. 2) Target Fusing. Instead of comparing features between targets, our FuseTracker fuses targets by refining their bounding boxes and merging the refined targets. These operations are conducted by a single DNN module, FuseTracker, such that targets from different sources~(tracking and detection) can be associated based on the refined bounding boxes and confidence scores. Similar to the post-processing of standard object detectors, FuseTracker uses the efficient non-max-suppression~(NMS) algorithm to produce the fused the targets.   

Figure~\ref{FigFFTNetwork} shows the FFT architecture that enclose FlowTracker and FuseTracker. The details of FlowTracker and FuseTracker are introduced below.

\begin{figure}[t]
\centering
\includegraphics[width=\linewidth]{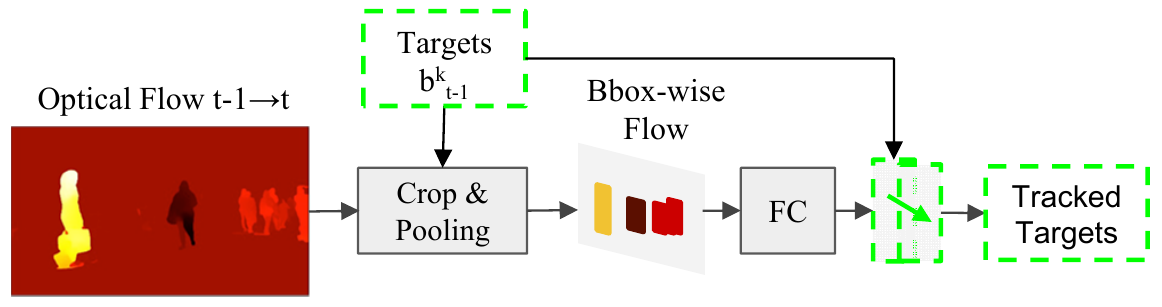}\hspace{-.0cm}
\caption{Network architecture of our FlowTracker.}
\label{FigFlowTracker}
\end{figure}

\subsection{FlowTracker Module}

The FlowTracker DNN module aims to estimate indefinite number of target-wise motion jointly from pixel-level features, \emph{i.e.}, optical flow. The network architecture of our FlowTracker is shown in Figure~\ref{FigFlowTracker}, in which we just take several convolutional layers and one fully-connected~(FC) layer to estimate the target motion from the previous frame to the current frame. Once our FlowTracker is trained, it just takes two frames as input and outputs multiple target motions between the two input frames.

The previous frame is denoted as $\mathbf{I}_{t-1}\in \mathbb{R}^{3 \times w \times h}$, and the current frame is represented by $\mathbf{I}_{t}\in \mathbb{R}^{ 3 \times w \times h}$, where $t > 1$ indicates the time index. Firstly, we feed the two frames into the FlowNet, and a pixel-wise optical flow $\mathbf{F}_t \in \mathbb{R}^{2 \times w \times h}$ can be obtained at the output layer. Secondly, we take the resulting optical flow into our FlowTracker, and train it using the ground truth motion vectors. In practice, the previous frame usually has multiple targets to be tracked in the current frame. For simplicity, we denote targets in the previous frame as $\mathbf{B}_{t-1}=\{\mathbf{b}_{t-1}^{k1}, \mathbf{b}_{t-1}^{k2}, \ldots, \mathbf{b}_{t-1}^{kn} | 1 \leq k_1, k_2,\ldots ,k_n \leq N\}$, where $\mathbf{b}_{t-1}^k = (x_{t-1}^k, y_{t-1}^k, w_{t-1}^k, h_{t-1}^k)$ indicates the position of $k\in\{k_1,k_2,\ldots,k_n\}$ target in frame $\mathbf{I}_{t-1}$. The corresponding targets in the current frame are denoted by $\mathbf{B}_t=\{\mathbf{b}_{t}^{k1}, \mathbf{b}_{t}^{k2}, \ldots, \mathbf{b}_{t}^{kn} | 1 \leq k_1, k_2,\ldots ,k_n \leq N\}$, where $\mathbf{b}_t^k = (x_t^k, y_t^k, w_t^k, h_t^k)$ indicates the position of $k\in\{k_1,k_2,\ldots,k_n\}$ target in frame $\mathbf{I}_t$. As a result, the target motions can be easily learned by measuring the difference between $\mathbf{B}_{t-1}$ and $\mathbf{B}_t$. The loss function for target motion is:
\begin{equation}
\label{eq_1}
L_1(\Delta \mathbf{B}_t,\Delta \mathbf{B}_t^*) =\|\Delta \mathbf{B}_t - \Delta \mathbf{B}_t^*\|_F^2.
\end{equation}
where $\Delta \mathbf{B}_t^* = \mathbf{B}_t - \mathbf{B}_{t-1}$ indicates the ground truth motion vectors of all targets between frame $\mathbf{I}_{t-1}$ and $\mathbf{I}_t$, and $\Delta \mathbf{B}_t$ denotes the output motion vectors by our FlowTracker DNN module. By optimizing this objective function to update the parameters of FlowTracker DNN module in the training process, our FlowTracker gains ability to predict the target motions $\Delta \mathbf{B}_t = \{[\Delta x_t^{k},\Delta y_t^{k}, \Delta w_t^{k}, \Delta h_t^{k}]\}$ from frame $\mathbf{I}_{t-1}$ to $\mathbf{I}_t$. As a result, we can easily estimate the target positions in the current frame as $\mathbf{B}_t = \mathbf{B}_{t-1} + \Delta \mathbf{B}_t$. When more and more target motions are consistently learned in the training process, our FlowTracker become stronger to associate targets across the adjacent frames.

\subsection{FuseTracker Module}

\begin{figure}[t]
\centering
\includegraphics[width=\linewidth]{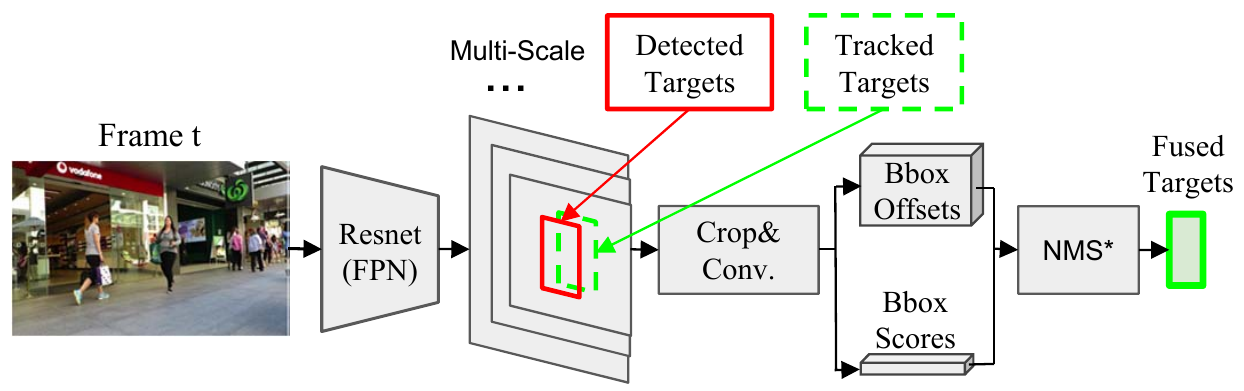}\hspace{-.0cm}
\caption{Network architecture of our FuseTracker.}
\label{FigFuseTracker}
\end{figure}

The FuseTracker DNN module aims to jointly refine and fuse targets tracked from FlowTracker $\mathbf{B}_t=\{b_{t}^{k_1},b_{t}^{k_2},\ldots ,b_{t}^{k_n}| 1 \leq k_1, k_2,\ldots ,k_n \leq N\}$ and the target detections $\mathbf{D}_t=\{d_{t}^{i}\}|_{i=1,\ldots,m}$ from image-based object detectors. The network architecture of FuseTracker is shown in Figure~\ref{FigFuseTracker}. Our FuseTracker is modified from the well-known Faster-RCNN~\cite{Ren_He_Girshick:2015}. Specifically, we add the Feature Pyramid Networks~(FPN)~\cite{LinDGHHB17} to extract features from different scales. In testing, we remove the Region Proposal Network~(RPN), and use the FlowTracker targets and public detection bboxes as proposals.

Neither the targets from FlowTracker nor the objects from image-based detector can be directly used as targets passing along to the next frames. On one hand, targets from our FlowTracker contains identity~(ID) index $\{k_1, k_2,\ldots ,k_n|1 \leq k_1, k_2,\ldots ,k_n \leq N\}$ inherited from the $\bf{I}_{t-1}$, but may have inaccurate bboxes in frame $t$. On the other hand, the objects from image-based detector require ID assignment, either taking the IDs of FlowTracker targets or generating new IDs. Moreover, even if some FlowTracker targets and detected objects can be matched, it is still unclear which bboxes should be used for targets in frame $t$. 

FuseTracker resolves this issue by treating both FlowTracker targets $\mathbf{B}_t$ and detected objects $\mathbf{D}_t$ equally as proposals. Formally, FuseTracker takes bbox proposals from both $\mathbf{B}_t$ and $\mathbf{D}_t$, and current frame $\mathbf{I}_{t}$ as input. For each proposal, FuseTracker estimates its offset $\Delta (cx,cy,w,h)$ and confidence score $c$. Given the two refined bbox sets $\mathbf{B}_t^{ref}$ and $\mathbf{D}_t^{ref}$, FuseTracker firstly deals with disappeared objects by killing the bboxes with low confidence scores. Then two-level NMS~(shown as ``NMS\textbf{*}" in Figure~\ref{FigFuseTracker}) are applied to merge the bboxes. In particular, the first level NMS is applied within the refined FlowTracker targets and detected objects to handle the object mutual occlusion. The second level NMS is conducted with Intersection over Union~(IoU) based bbox matching and max-score selection. This level of NMS assigns IDs from FlowTracker targets to detected objects, and produces more accurate bboxes for the fused targets. All the detected objects that do not inherit target IDs are treated as new targets.

In training FuseTracker, similar to Faster-RCNN~\cite{Ren_He_Girshick:2015}, the RPN generates multiple scales of bounding box proposals, and the feature maps are cropped by each proposal. Then the Region of Interest~(RoI) pooling is applied to the cropped feature maps, which are later passed to both bbox classification head and regression head to get object classification score and bbox location offsets, respectively. Given the predicted classification score, location offset and ground truth bounding boxes, we minimize an objective function following the multi-task loss~\cite{Girshick:2015:FR:2919332.2920125}. The loss function for the detection head is defined as:

\begin{equation}
\label{eq_3}
L_2(c,c^*,b,b^*) = L_{cls}(c,c^*) + \lambda [c^*\geq1] L_{reg}(b,b^*),
\end{equation}
where $c$ and $b=(x,y,w,h)$ are the predicted classification score and bounding box location, respectively. $c^*$ and $b^*$ are the corresponding ground truth. The classification loss $L_{cls}$ is log loss and the regression loss $L_{reg}$ is the smooth L1 loss. Particularly, the weights for the RPN part are fixed during training since we just want to use the RPN to produce bbox proposals which imitates the FlowTracker output and public detection results.

In the testing stage of FuseTracker, we remove the RPN, and use the FlowTracker targets $\mathbf{B}_t$ and bounding boxes from public detectors $\mathbf{D}_t$ as proposals.

\subsection{Inference algorithm}
\label{inferalgorithm}

The FFT inference algorithm runs the FlowTracker and FuseTracker jointly to produce MOT outputs (see details in Algorithm \ref{algorithm1}). In summary, there are $7$ steps in the FFT inference algorithm:
\begin{itemize}
  \setlength\itemsep{-0.1em}
  \item \textbf{Step 1}. The detections in the first frame $\mathbf{D}_0$ is refined by the FuseTracker. The detected bboxes are passed to FuseTracker as the proposals. Among the output bboxes of FuseTracker, those whose confidence scores are smaller than $\mathrm{thresh\_score}$ are killed and frame-wised NMS is applied. The refined detection $\mathbf{D}_0^{ref}$ is used to initialize the trajectory set $\mathbf{T}$. (Line 2-7)
  \item \textbf{Step 2}. The detections for the current frame $\mathbf{D}_{t}$ is refined by the FuseTracker in the same way as \textbf{Step 1}. (Line 8)
  \item \textbf{Step 3}. All the tracked bboxes $\mathbf{B}_{t-1}$ (bboxes with trajectory IDs) in previous frame $\mathbf{I}_{t-1}$ are found in $\mathbf{T}$. The FlowTracker takes the image pair $\{\mathbf{I}_{t-1}, \mathbf{I}_{t}\}$ and $\mathbf{B}_{t-1}$ as input, and produces the corresponding tracked bboxes $\mathbf{B}_{t}$ in the current frame $\mathbf{I}_{t}$. (Line 9-11)
  \item \textbf{Step 4}. The FuseTracker fuses the bboxes from $\mathbf{B}_{t}$ and $\mathbf{D}_{t}^{ref}$. We first refine the $\mathbf{B}_{t}$ the same way as described in \textbf{Step 1}. Then, for each tracked bbox $\mathbf{b}_{t}^k$ in $\mathbf{B}_{t}^{ref}$, we find the bbox that has the largest IoU from $\mathbf{D}_{t}^{ref}$. If this IoU is larger than $\mathrm{thresh\_iou}$, we consider this detected bbox is matched to $\mathbf{b}_{t}^k$. If the confidence score for the matched detected bbox is higher than $\mathbf{b}_{t}^k$'s score, we use the detected bbox to replace $\mathbf{b}_{t}^k$. Besides, the NMS is applied to the fused bbox set to get the tracked bbox result $\mathbf{B}_{t}^{trk}$. The FuseTracker returns the NMS output $\mathbf{B}_{t}^{trk}$ and the unmatched detections $\mathbf{D}_{t}^{umt}$. (Line 12)
  \item \textbf{Step 5}. The bboxes in $\mathbf{B}_{t}^{trk}$ are add to the corresponding trajectories in $\mathbf{T}$. (Line 13)
  \item \textbf{Step 6}. For the unmatched detections $\mathbf{D}_{t}^{umt}$, we conduct backtracking~(see more details below) to match them with trajectories $\mathbf{T}_{nas}$ (trajectories not successfully associated in the current frame by \textbf{Step 2-5}). (Line 14-19)
  \item \textbf{Step 7}. The detections that cannot be matched by backtracking $\mathbf{D}_{t}^{umt'}$ are initialized as new trajectories. (Line 20)
\end{itemize}

\textbf{Backtracking}. In the tracking process, some targets are occluded or out-of-scene in the previous frames and re-emerge in the current frame. To handle these cases, one needs to track repeatedly between the current frame and several previous frames. We call this process BackTracking~(BT). To be specific, for the trajectories $\mathbf{T}_{nas}$, we try to find bboxes of these trajectories $\mathbf{B}_{nas}$ for the earlier frames. We apply the FlowTracker to shift bboxes in $\mathbf{B}_{nas}$ from previous frames to $\mathbf{B}_{t}'$ in the current frame. Then the FuseTracker is applied to fuse the unmatched detections $\mathbf{D}_{t}^{umt}$ and $\mathbf{B}_{t}'$. Here, we only consider the trajectories that could be matched with the detections. Finally, we add the tracked bboxes $\mathbf{B}_{t}^{trk'}$ to the corresponding trajectories. The more previous frames used in BT, the more likely all targets can be tracked despite of occlusions. We study the effects of the number of BT frames in the ablation studies section below. However, the inference overheads increase with the increase of BT frames. For efficient implementation, we apply different numbers of BTs adaptively at different frames and process multiple BTs in one batch.

\begin{algorithm}
	\caption{Flow-Fuse-Tracker} 
	\renewcommand{\algorithmicrequire}{\textbf{Input:}}  
         \renewcommand{\algorithmicensure}{\textbf{Output:}} 
	\begin{algorithmic}[1]
	     \Require Video sequence $\mathbf{I} = \{\mathbf{I}_{t}\}_{t=0}^{T-1}$ of frame $\mathbf{I}_{t}$ at time $t$; Public detection set $\mathbf{D}=\{\mathbf{D}_{t}\}_{t=0}^{T-1}$ of detections $\mathbf{D}_{t}$ for frame $\mathbf{I}_{t}$
	     \Ensure Trajectory set $\mathbf{T}=\{\mathbf{T}^{k}\}_{k=1}^{M}$, where $\mathbf{T}^{k}=\{\mathbf{b}_{t}^{k}\}_{t=0}^{T-1}$ with $\mathbf{b}_{t}^{k}=(x,y,w,h,t)$
	     \State $\mathbf{T} \leftarrow \O$
		\For {$\mathbf{I}_{t}, \mathbf{D}_{t}$ $in$ $\mathrm{zip}(\mathbf{I}, \mathbf{D})$}
			\If {$t==0$} 
				\State $\mathbf{D}_{0}^{ref} \leftarrow \mathrm{FuseTracker}(\mathbf{I}_{0}, \mathbf{D}_{0})$
				\State Initialize the trajectory set $\mathbf{T}$ with $\mathbf{D}_{0}^{ref}$
				\State \textbf{continue}
			\EndIf
			\State $\mathbf{D}_{t}^{ref} \leftarrow \mathrm{FuseTracker}(\mathbf{I}_{t},\mathbf{D}_{t})$ 
			\State $\mathbf{B}_{t-1} \leftarrow$ find all tracked $\mathbf{b}_{t-1}^{k}$ at $\mathbf{I}_t$ from $\mathbf{T}$ 
			\State $\mathbf{I}_{pair} \leftarrow \{\mathbf{I}_{t-1}, \mathbf{I}_{t}\}$
			\State $\mathbf{B}_{t} \leftarrow \mathrm{FlowTracker}(\mathbf{I}_{pair}, \mathbf{B}_{t-1})$
			\State $\mathbf{B}_{t}^{trk}, \mathbf{D}_{t}^{umt} \leftarrow \mathrm{FuseTracker}(\mathbf{B}_{t}, \mathbf{D}_{t}^{ref})$ 
			\State $\mathbf{T} \leftarrow$ update $\mathbf{B}_{t}^{trk}$ to the corresponding $\mathbf{T}^{k}$ 
			\State $\mathbf{T}_{nas} \leftarrow$ find tracks $\mathbf{T}^{k}$ that do not have $\mathbf{b}_{t}^{k}$ 
			\State $\mathbf{I}_{pairs} {\hspace{-0.2em}} \leftarrow$ get image pairs for back tracking, $\mathbf{I}_{pairs}= \{\{\mathbf{I}_{t}, \mathbf{I}_{t-2}\}, \{\mathbf{I}_{t}, \mathbf{I}_{t-3}\},\ldots ,\{\mathbf{I}_{t}, \mathbf{I}_{t-\mathrm{BT}}\}\}$
			\State $\mathbf{B}_{nas} \leftarrow$ get tracked bboxes from $\mathbf{T}_{nas}$ for back tracking, $\mathbf{B}_{nas} = \{\mathbf{B}_{t-2}, \mathbf{B}_{t-3},\ldots, \mathbf{B}_{t-\mathrm{BT}}\}$
			\State $\mathbf{B}_{t}^{'} \leftarrow \mathrm{FlowTracker}(\mathbf{I}_{pairs}, \mathbf{B}_{nas})$
			\State $\mathbf{B}_{t}^{trk'} , \mathbf{D}_{t}^{umt'} \leftarrow \mathrm{FuseTracker}(\mathbf{B}_{t}^{'}, \mathbf{D}_{t}^{umt})$
			\State $\mathbf{T} \leftarrow $ update $\mathbf{B}_{t}^{trk'}$ to the corresponding $\mathbf{T}^{k}$
			\State $\mathbf{T} \leftarrow$ initialize new trajectories with $\mathbf{D}_{t}^{umt'}$ 
		\EndFor
		\State return $\mathbf{T}$
	\end{algorithmic} 
	\label{algorithm1}
\end{algorithm}

\section{Experiments}
FFT is implemented in Python with PyTorch framework. Extensive experiments have been conducted on the MOT benchmarks, including the 2DMOT2015~\cite{MOTChallenge2015}, MOT16~\cite{MOT16}, and MOT17~\cite{MOT16}. In particular, we compare FFT with the latest published MOT approaches, perform ablation studies on the effectiveness of FuseTracker, FlowTracker and BackTracking frames, and also analyze the influence of target sizes and percentages of occlusion on the performance of FFT. 

\begin{table*}[h]
\footnotesize
\centering 
\setlength{\tabcolsep}{3pt}
\begin{tabular}{c p{3cm}<{\centering} c c c c c c c c c c c}
\hline 
& Method & MOTA $\uparrow$ & MOTP $\uparrow$ & IDF1 $\uparrow$ & IDP $\uparrow$ & IDR $\uparrow$ & MT $\uparrow$& ML$\downarrow$ & FP $\downarrow$ & FN $\downarrow$ & ID Sw. $\downarrow$ & Frag. $\downarrow$ \\ [0.5ex]  
\hline\hline 
& FFT w/o FuseTracker & 50.1 & 75.7 & 44.2 &  58.5 & 35.6 & 18.8\% & 33.1\% & 27,213 & 248,011 & 6,429 & 6,956 \\
\hline
& FFT w/o FlowTracker & 55.8 & 77.5  & 49.3 & 60.8 & 41.5 & 26.1\% & 27.1\% & 31,172 & \textbf{210,196} & 7,866 & 5,519 \\
\hline
& FFT &\textbf{56.5} & \textbf{77.8} & \textbf{51.0} & \textbf{64.1} & \textbf{42.3} & \textbf{26.2\%} & \textbf{26.7\%} & \textbf{23,746} & 215,971 & \textbf{5,672} & \textbf{5,474} \\
\hline\\
\end{tabular}
\caption{Ablation study of FFT on the effectiveness of FuseTracker and FlowTracker on the MOT17 benchmark. \textbf{(Best in bold)}} \label{tab:ablationFlowTracker}
\end{table*}

\begin{table*}[h]
\footnotesize
\centering 
\setlength{\tabcolsep}{3pt}
\begin{tabular}{c p{3cm}<{\centering} c c c c c c c c c c c c}
\hline 
& Ablation & MOTA $\uparrow$ & MOTP $\uparrow$ & IDF1 $\uparrow$ & IDP $\uparrow$ & IDR $\uparrow$ & MT $\uparrow$& ML$\downarrow$ & FP $\downarrow$ & FN $\downarrow$ & ID Sw. $\downarrow$ & Frag. $\downarrow$ \\ [0.5ex]  
\hline\hline 
& FFT-BT1 & 55.9 & \textbf{77.8} & 40.2 & 50.6 & 33.4 & \textbf{26.3\%} & \textbf{26.7\%} & \textbf{23,540} & \textbf{215,740} & 9,789 & 5,495  \\
& FFT-BT10 & 56.4 & \textbf{77.8} & 49.1 & 61.8 & 40.7 & 26.2\% & \textbf{26.7\%} & 23,683 & 215,941 & 6,227 & 5,479 \\
& FFT-BT20 & \textbf{56.5} &  \textbf{77.8} & 50.3 & 63.3 & 41.7 & 26.2\% & \textbf{26.7\%} & 23,743 & 216,000 & 5,810 & 5,484  \\
& FFT-BT30 &\textbf{56.5} & \textbf{77.8} & \textbf{51.0} & \textbf{64.1} & \textbf{42.3} & 26.2\% & \textbf{26.7\%} & 23,746 & 215,971 & \textbf{5,672} & \textbf{5,474}  \\
\hline\\
\end{tabular}
\caption{ Ablation study of FFT on the number of Backtracking frames (BT\#) on the MOT17 benchmark. \textbf{(Best in bold)}} \label{tab:ablationBT}
\end{table*}

\subsection{Experiment Setting}
\textbf{Training.} FlowTracker consists of a FlowNet2~\cite{IMSKDB17} component and a regression component. The FlowNet2 is pre-trained on MPI-Sintel~\cite{Butler:ECCV:2012} dataset and its weights are fixed during the training of the regression component. In the training process of our FlowTracker, we use the Adam optimizer and set the initial learning rate to 1e-4, divided by 2 at the epoch 80, 160, 210 and 260, respectively. We train the model for 300 epochs with a batch size of 24.  

FuseTracker is modified from the Faster-RCNN, in which we take ResNet101+FPN as the backbone. The backbone network is initialized with the ImageNet pre-trained weights and the whole model is trained with the ground truth bounding boxes from the MOT datasets. In the training process of our FuseTracker, we use the Stochastic Gradient Descent~(SGD) optimizer and set the initial learning rate to 1e-4, divided by 10 every 5 epochs. We train the model for 20 epochs with a batch size of 16.

\textbf{Inference.}
As described in \ref{inferalgorithm}, the FFT inference algorithm is specified by three parameters: the confidence score threshold $\mathrm{thresh\_score}$, the IoU threshold $\mathrm{thres\_iou}$, the NMS threshold $\mathrm{thresh\_nms}$, and the number of backtracking frames. In particular, we use $\mathrm{thresh\_score}$ = 0.5, $\mathrm{thresh\_iou}$ = 0.5 and $\mathrm{thresh\_nms}$ = 0.5 for all the benchmark datasets in our experiment. We use different backtracking frames for different datasets due to their frame rate difference. The settings are specified in \ref{evaluation}. Also the effects of BT frames were studied in \ref{ablation}.

\subsection{Main Results}
\label{evaluation}
\textbf{Datasets}.
We use three benchmark datasets, including the 2DMOT2015~\cite{MOTChallenge2015}, MOT16~\cite{MOT16}, and MOT17~\cite{MOT16}, to conduct our experiments. All of them contain video sequences captured by both static and moving cameras in unconstrained environments, \textit{e.g.}, under the complex scenes of illumination changes, varying viewpoints and weather conditions. The 2DMOT2015 benchmark contains 22 video sequences, among which 11 videos sequences are for testing. The 2DMOT2015 benchmark provides ACF~\cite{Dollar:2014:FFP:2693345.2693405} detection results. The Frames Per Second (FPS) for each video in 2DMOT2015 varies from 2.5 to 30. Therefore we set our BT frames according to the FPS. We used BT frames 3 for low-FPS videos of 2.5, BT frames 10 for medium-FPS videos of 7 and 10, and BT frames 30 for high-FPS videos of 14, 25, and 30. The MOT16 benchmark includes $7$ fully annotated training videos and $7$ testing videos. The public detection results are obtained by DPM~\cite{Felzenszwalb_Girshick_McAllester:2010}. We use BT frames 30 for all the testing videos. The MOT17 benchmark is consist of  $7$ fully annotated training videos and $7$ testing videos, in which public detection results are obtained by three image-based object detectors: DPM~\cite{Felzenszwalb_Girshick_McAllester:2010}, FRCNN~\cite{Ren_He_Girshick:2015} and SDP~\cite{Ess_Leibe_Van:2007}. The BT frames for all the testing videos are set to be 30.

\begin{figure}[t]
\centering
\includegraphics[width=\linewidth]{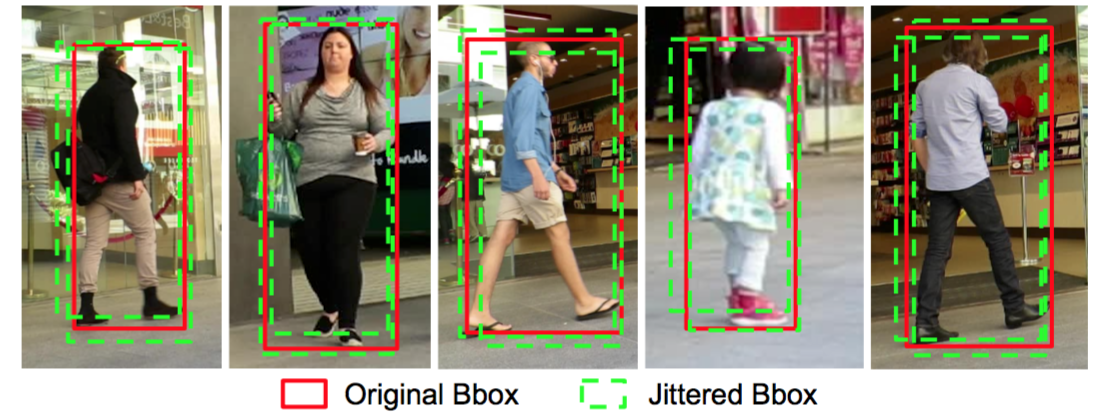}\hspace{-.0cm}
\caption{The jittering of the bounding boxes. The \textcolor{red}{red} bboxes are the ground truth bboxes. We apply the random jittering to them to slightly shift their positions and change their shapes, and we use the jittered bboxes (\textcolor{green}{green} dashed bboxes) in training the FlowTracker.}
\label{jittering}
\end{figure}

\textbf{Augmentation.} 
We apply two types of augmentation methods for training the FlowTracker. In order to deal with the different movement speeds of multiple target objects and the different BT frames, we use different sampling rates to get the image pairs. Particularly, each video sequence is processed by 8 different sampling rates, which results in 8 sets. The intervals between the image frame numbers in each image pair for the 8 sets are 1, 3, 5, 10, 15, 20, 25, 30, respectively. Moreover, when we want to get target motion out of the pixel-wise optical flow, if the bbox location in the current frame is not accurate, the cropped flow feature will be incomplete but we still want to regress to the accurate bbox location in the next frame. Therefore, as shown in Figure \ref{jittering}, we apply jittering to the bboxes for regression. We enlarge or reduce the bbox height and width by a random ratio between 0 and 0.15. Also, we shift the center of the bbox vertically and horizontally by a random ratio between -0.15 and 0.15 of the width and height, respectively. We ensure the IoU between the original bbox and the bbox after jittering is larger than 0.8.
To train the FuseTracker, we just use horizontal flip as the augmentation.

\begin{table*}[t]
\footnotesize
\centering 
\setlength{\tabcolsep}{3pt}
\begin{tabular}{c p{2.4cm}<{\centering} c c c c c c c c c c c}
\hline 
& Method & MOTA $\uparrow$ & MOTP $\uparrow$ & IDF1 $\uparrow$ & IDP $\uparrow$ & IDR $\uparrow$ & MT $\uparrow$ & ML $\downarrow$ & FP $\downarrow$ & FN $\downarrow$ & ID Sw. $\downarrow$ & Frag. $\downarrow$ \\[0.5ex]  
\hline\hline 
\multirow{1}{*}{Offline} 
& JointMC~\cite{keuper2018motion} & 35.6 & 71.9 & 45.1 & 54.4 & 38.5 & 23.2\% & 39.3\% & 10,580 & 28,508 & 457 & \textbf{969}  \\
\hline
\multirow{10}{*}{Online} 
& INARLA~\cite{wu2019instance} & 34.7 & 70.7 & 42.1 & 51.7 & 35.4 & 12.5\% & 30.0\% & 9,855 & 29,158 & 1,112 & 2,848 \\
& HybridDAT~\cite{yang2017hybrid} & 35.0 & 72.6 & 47.7 & 61.7 & 38.9 & 11.4\% & 42.2\% & 8,455 & 31,140 & \textbf{358} & 1,267 \\
& RAR15pub~\cite{fang2018recurrent} & 35.1 & 70.9 & 45.4 & 62.0 & 35.8 & 13.0\% & 42.3\% & 6,771 & 32,717 & 381 & 1,523 \\
& AMIR15~\cite{Sadeghian_Alahi_Savarese:2017} & 37.6 & 71.7 & 46.0 & 58.4 & 38.0 & 15.8\% & 26.8\% & 7,933 & 29,397 & 1,026 & 2,024 \\
& STRN~\cite{xu2019spatial} & 38.1 & 72.1 & 46.6 & 63.9 & 36.7 & 11.5\% & 33.4\% & 5,451 & 31,571 & 1,033 & 2,665 \\
& AP\_HWDPL\_p~\cite{Chen_Ai_Shang:2017} & 38.5 & 72.6 & 47.1 & \textbf{68.4} & 35.9 & 8.7\% & 37.4\% & \textbf{4,005} & 33,203 & 586 & 1,263 \\
& KCF~\cite{chu2019online} & 38.9 & 70.6 & 44.5 & 57.1 & 36.5 & 16.6\% & 31.5\% & 7,321 & 29,501 & 720 & 1,440 \\
& Tracktor~\cite{bergmann2019tracking} & 44.1 & 75.0 & 46.7 & 58.1 & 39.1 & 18.0\% & 26.2\% & 6,477 & 26,577 & 1,318 & 1,790 \\
& \textbf{FFT(Ours)} & \textbf{46.3} & \textbf{75.5} & \textbf{48.8} & 54.8 & \textbf{44.0} & \textbf{29.1\%} & \textbf{23.2\%} & 9,870 & \textbf{21,913} & 1,232 & 1,638\\%
\hline\\
\end{tabular}
\caption{Comparison between FFT and other modern MOT methods on the 2DMOT15 benchmark. \textbf{(Best in bold)}} \label{tab:mot15results}
\vspace{-0.2cm}
\end{table*}

\begin{figure}[t]
\centering
\includegraphics[width=\linewidth]{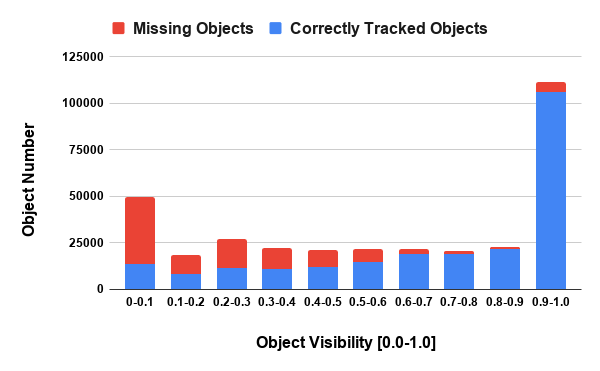}\hspace{-.0cm}
\includegraphics[width=\linewidth]{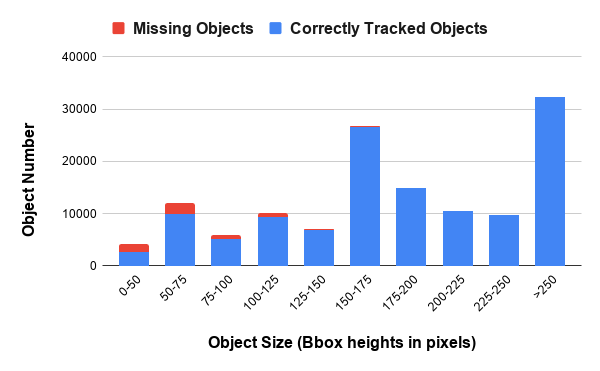}\hspace{-.0cm}
\caption{Effects of object visibility and object sizes on FFT (MOT17 training set).}
\label{FigAnalysis}
\vspace{-0.15cm}
\end{figure}

\textbf{Results}.
To evaluate the performance of our proposed method, we follow the CLEAR MOT Metrics~\cite{Bernardin2008}. We report the Multiple Object Tracking Precision (MOTP) and the Multiple Object Tracking Accuracy (MOTA) that combines three sources of errors: False Positives~(FP), False Negatives~(FN) and the IDentity Switches~(IDS). Additionally, we also report the ID F1 Score~(IDF1), ID Precision~(IDP), ID Recall~(IDR), the number of track fragmentations~(Frag.), the percentage of mostly tracked targets~(MT) and the percentage of mostly lost targets~(ML). Among them, \textbf{the MOTA score is the main criterium}~\footnote{~\url{https://motchallenge.net/workshops/bmtt2019/}}.

We compare FFT method with the latest MOT approaches we can find based on their published papers\footnote{For MOT leaderboard submissions that do not include published or Arxiv papers, we could not analyze their methods and validate their effectiveness}. Table~\ref{tab:mot15results}, \ref{tab:mot16results} and \ref{tab:mot17results} list the MOT results of two main categories: offline trackers and online trackers. Note that offline trackers typically use information from both previous and future frames in tracking targets in the current frame, whereas online trackers only use knowledge from previous frames. Therefore, in terms of the main MOT metric, MOTA, offline trackers generally perform better than the online ones. However, FFT, which is an online method, outperforms all the MOT approaches in 2DMOT15, MOT16 and MOT17 benchmark in terms of MOTA as shown in Table~\ref{tab:mot15results}, \ref{tab:mot16results} and \ref{tab:mot17results}. Moreover, FFT also achieves the best performance in the mostly tracked targets (MT), the mostly lost targets (ML) and false negative (FN) metrics. It indicates that FFT could correctly track the most positions. 

\begin{table*}[t]
\footnotesize
\centering 
\setlength{\tabcolsep}{3pt}
\begin{tabular}{c p{2.4cm}<{\centering} c c c c c c c c c c c}
\hline 
& Method & MOTA $\uparrow$ & MOTP $\uparrow$ & IDF1 $\uparrow$ & IDP $\uparrow$ & IDR $\uparrow$ & MT $\uparrow$ & ML $\downarrow$ & FP $\downarrow$ & FN $\downarrow$ & ID Sw. $\downarrow$ & Frag. $\downarrow$ \\[0.5ex]  
\hline\hline 
\multirow{8}{*}{Offline} 
& FWT~\cite{henschel2019multiple} & 47.8 & 75.5 & 44.3 & 60.3 & 35.0 & 19.1\% & 38.2\% & 8,886 & 85,487 & 852 & 1,534 \\
& GCRA~\cite{ma2018trajectory} & 48.2 & 77.5 & 48.6 & 69.1 & 37.4 & 12.9\% & 41.1\% & 5,104 & 88,586 & 821 & 1,117 \\
& TLMHT~\cite{sheng2018iterative} & 48.7 & 76.4 & 55.3 & 76.8 & 43.2 & 15.7\% & 44.5\% & 6,632 & 86,504 & \textbf{413} & 642 \\
& LMP~\cite{Tang_Andriluka_Andres:2017} & 48.8 & \textbf{79.0} & 51.3 & 71.1 & 40.1 & 18.2\% & 40.1\% & 6,654 & 86,245 & 481 & 595 \\
& AFN~\cite{shen2018tracklet} & 49.0 & 78.0 & 48.2 & 64.3 & 38.6 & 19.1\% & 35.7\% & 9,508 & 82,506 & 899 & 1,383 \\
& eTC~\cite{wang2018exploit} & 49.2 & 75.5 & 56.1 & 75.9 & \textbf{44.5} & 17.3\% & 40.3\% & 8,400 & 83,702 & 606 & 882 \\
& HCC~\cite{ma2018customized} & 49.3 & \textbf{79.0} & 50.7 & 71.1 & 39.4& 17.8\% & 39.9\% & 5,333 & 86,795 & 391 & \textbf{535} \\
& NOTA~\cite{chen2019aggregate} & 49.8 & 74.5 & 55.3 & 75.3 & 43.7 & 17.9\% & 37.7\% & 7,248 & 83,614 & 614 & 1,372 \\
\hline
\multirow{7}{*}{Online} 
& MOTDT~\cite{chen2018real} & 47.6 & 74.8 & 50.9 & 69.2 & 40.3 & 15.2\% & 38.3\% & 9,253 & 85,431 & 792 & 1,858 \\
& STRN~\cite{xu2019spatial} & 48.5 & 73.7 & 53.9 & 72.8 & 42.8 & 17.0\% & 34.9\% & 9,038 & 84,178 & 747 & 2,919 \\
& KCF~\cite{chu2019online} & 48.8 & 75.7 & 47.2 & 65.9 & 36.7 & 15.8\% & 38.1\% & 5,875 & 86,567 & 906 & 1,116 \\ 
& LSSTO~\cite{feng2019multi} & 49.2 & 74.0 & \textbf{56.5} & \textbf{77.5} & \textbf{44.5} & 13.4\% & 41.4\% & 7,187 & 84,875 & 606 & 2,497 \\
& Tracktor~\cite{bergmann2019tracking} & 54.4 & 78.2 & 52.5 & 71.3 & 41.6 & 19.0\% & 36.9\% & \textbf{3,280} & 79,149 & 682 & 1,480 \\
& \textbf{FFT(Ours)} & \textbf{56.5} & 78.1 & 50.1 & 64.4 & 41.1 & \textbf{23.6\%} & \textbf{29.4\%} & 5,831 & \textbf{71,825} & 1,635 & 1,607 \\%
\hline\\
\end{tabular}
\caption{Comparison between FFT and other modern MOT methods on the MOT16 benchmark. \textbf{(Best in bold)}} \label{tab:mot16results}
\vspace{-0.2cm}
\end{table*}

\begin{table*}[h]
\footnotesize
\centering 
\setlength{\tabcolsep}{3pt}
\begin{tabular}{c p{2.4cm}<{\centering} c c c c c c c c c c c}
\hline 
& Method & MOTA $\uparrow$ & MOTP $\uparrow$ & IDF1 $\uparrow$ & IDP $\uparrow$ & IDR $\uparrow$ & MT $\uparrow$ & ML $\downarrow$ & FP $\downarrow$ & FN $\downarrow$ & ID Sw. $\downarrow$ & Frag. $\downarrow$ \\[0.5ex]  
\hline\hline 
\multirow{8}{*}{Offline} 
& jCC~\cite{keuper2018motion} & 51.2 & 75.9 & 54.5 & 72.2 & 43.8 & 20.9\% & 37.0\% & 25,937 & 247,822 & 1,802 & 2,984 \\
& NOTA~\cite{chen2019aggregate} & 51.3 & 76.7 & 54.5 & 73.5 & 43.2 & 17.1\% & 35.4\% & 20,148 & 252,531 & 2,285 & 5,798 \\
& FWT~\cite{henschel2019multiple} & 51.3 & 77.0 & 47.6 & 63.2 & 38.1 & 21.4\% & 35.2\% & 24,101 & 247,921 & 2,648 & 4,279 \\
& AFN17~\cite{shen2018tracklet} & 51.5 & 77.6 & 46.9 & 62.6 & 37.5 & 20.6\% & 35.5\% & 22,391 & 248,420 & 2,593 & 4,308 \\
& eHAF~\cite{sheng2018heterogeneous} & 51.8 & 77.0 & 54.7 & 70.2 & 44.8 & 23.4\% & 37.9\% & 33,212 & 236,772 & 1,834 & \textbf{2,739} \\
& eTC~\cite{wang2018exploit} & 51.9 & 76.3 & 58.1 & 73.7 & 48.0 & 23.1\% & 35.5\% & 36,164 & 232,783 & 2,288 & 3,071 \\
& JBNOT~\cite{henschel2019multiple} & 52.6 & 77.1 & 50.8 & 64.8 & 41.7 & 19.7\% & 35.8\% & 31,572 & 232,659 & 3,050 & 3,792 \\
& LSST~\cite{feng2019multi} & 54.7 & 75.9 & \textbf{62.3} & \textbf{79.7} & \textbf{51.1} & 20.4\% & 40.1\% & 26,091 & 228,434 & \textbf{1,243} & 3,726 \\
\hline
\multirow{7}{*}{Online} 
& MOTDT~\cite{chen2018real} & 50.9 & 76.6 & 52.7 & 70.4 & 42.1 & 17.5\% & 35.7\% & 24,069 & 250,768 & 2,474 & 5,317 \\
& STRN~\cite{xu2019spatial} & 50.9 & 75.6 & 56.0 & 74.4 & 44.9 & 18.9\% & 33.8\% & 25,295 & 249,365 & 2,397 & 9,363 \\
& FAMNet~\cite{chu2019famnet} & 52.0 & 76.5 & 48.7 & 66.7 & 38.4 & 19.1\% & 33.4\% & 14,138 & 253,616 & 3,072 & 5,318 \\
& LSSTO~\cite{feng2019multi} & 52.7 & 76.2 & 57.9 & 76.3 & 46.6
 & 17.9\% & 36.6\% & 22,512 & 241,936 & 2,167 & 7,443 \\
& Tracktor~\cite{bergmann2019tracking} & 53.5 & \textbf{78.0} & 52.3 & 71.1 & 41.4 & 19.5\% & 36.6\% & \textbf{12,201}& 248,047 & 2,072 & 4,611 \\
& \textbf{FFT(Ours)} & \textbf{56.5} & 77.8 & 51.0 & 64.1 & 42.3 & \textbf{26.2\%} & \textbf{26.7\%} & 23,746 & \textbf{215,971} & 5,672 & 5,474\\ %
\hline\\
\end{tabular}
\caption{Comparison between FFT and other modern MOT methods on the MOT17 benchmark. \textbf{(Best in bold)}} \label{tab:mot17results}
\vspace{-0.22cm}
\end{table*}

\subsection{Ablation Study}
\label{ablation}
This section presents the ablation study on the MOT17 benchmark. We show the effectiveness of our FuseTracker and FlowTracker in Table~\ref{tab:ablationFlowTracker}. We first remove the FuseTracker to prove it is effective. However, we do not have any cues to generate scores for bboxes without using FuseTracker. In order to keep the completeness of our algorithm logic, we retrain a small RPN-like network with ResNet18 backbone to produce the scores for bboxes. Besides, we choose to trust public detections more here since we do not have the refinement module anymore. FFT w/o FuseTracker in Table~\ref{tab:ablationFlowTracker} shows that the MOTA improves by $6.4$ and the IDF1 improves by $6.8$ with FuseTracker. All the other metrics are also improved by including FuseTracker in FFT. Then, we show the FlowTracker is effective. By removing the FlowTracker, we directly use the bboxes in the previous frame as the tracked target proposals for the FuseTracker in the current frame. FFT w/o FlowTracker in Table~\ref{tab:ablationFlowTracker} shows that with FlowTracker the MOTA improves by $0.7$ and the IDF1 improves by $1.7$. Almost all the other metrics are also improved by including FlowTracker in FFT. 

We also study the effectiveness of BackTracking (BT). As shown in Table~\ref{tab:ablationBT}, BT highly affects the two main metrics MOTA and IDF1. We run FFT with fixed BT frame number of $1$, $10$, $20$, and $30$. As the BT frame number increases, the MOTA increases by $0.6$ to $56.5$ and the IDF1 increase by $10.8$ to $51.0$. The performance saturates at around BT30. The results show that with a bigger number of BT frames, FFT observes longer temporal clips and is more likely to track the occluded, out-of-scene and noisy objects.

\subsection{Analysis}
Finally we analyze the specific performance of FFT in more details. Since our FlowTracker computes pixel motion (optical flows) at the frame bases, its performance is likely to be affected by occlusion and object sizes. When an object is occluded in the current frame, optical flow may not match all pixels between frames. Moreover, computation of optical flow at the frame level seeks to average errors over all pixels, so the optical flow of small objects (fewer pixels) is more likely to be inaccurate, which will further affect the target-wise motion estimation. We show the effects of object visibility and object sizes in following experiments.

Since the ground truth of the MOT test sets is not publicly available, we perform our analysis on the MOT17 training set. Figure~\ref{FigAnalysis} reports the total number of correctly tracked objects and missing objects summed over three public object detectors of MOT17, with respect to the objects' visibility and size respectively. Object visibility of a target bbox is computed as the ratio between non-overlapped area divided by its bbox area. As shown in the subfigure~\textbf{Object Visibility}, at low visibility, more objects are missing. The percentage of correctly tracked objects rises as the visibility increases. After the object visibility reaches $0.8$, the correctly tracked object ratio becomes steady. 

For the object size analysis, we only collect the objects whose visibility is higher than $0.8$ in order to reduce the variables in experiments. The size of an object is measured by its bbox height. In the subfigure \textbf{Object Size}, the percentage of correctly tracked objects rises as the object size increase. For the object size (bbox height) reaches $150$, the tracking performance is nearly perfect.

\section{Conclusion}
\vspace{0.17cm}
Flow-Fuse-Tracker~(FFT) achieves new state-of-the-arts of MOT by addressing two fundamental issues in existing approaches: 1) how to share motion computation over indefinite number of targets. 2) how to fuse the tracked targets and detected objects. Our approach presents an end-to-end DNN solution to both issues and avoid pair-wise appearance-based Re-ID, which has been the costly and indispensable components of most existing MOT approaches. FFT is a very general MOT framework. The presented version can be further improved by using better optical flow networks in FlowTracker and more recent object detector structures in FuseTracker. 

{
\small
\bibliographystyle{ieee_fullname}
\bibliography{FlowNet}
}

\end{document}